\documentclass[letterpaper, 10 pt, conference]{ieeeconf}
\IEEEoverridecommandlockouts                            

\usepackage[utf8]{inputenc}
\usepackage[T1]{fontenc}
\usepackage{graphicx}
\usepackage{float} 
\usepackage{ragged2e}
\usepackage{booktabs}
\usepackage{bbding}
\usepackage{multirow}
\usepackage{makecell}
\usepackage{amsmath,amssymb,amsfonts}
\usepackage[table,xcdraw]{xcolor}
\usepackage{cite}
\usepackage[colorlinks,linkcolor=blue]{hyperref}
\usepackage{cleveref}
\usepackage{tabularx}
\usepackage[caption=true,font=normalsize,labelfont=sf,textfont=sf]{subfig}
\usepackage{textcomp}
\usepackage{stfloats}
\usepackage{tabularx}
\usepackage[linesnumbered,ruled,vlined]{algorithm2e}
\usepackage{adjustbox}

\hypersetup{
    colorlinks=true, 
    linkcolor=blue,  
    filecolor=blue,  
    urlcolor=black,  
    citecolor=blue, 
    pdfborder={0 0 0}
}

\Crefname{figure}{Fig.}{Figs.}
\Crefname{equation}{Eq.}{Eqs.}

\title{\bf Bridging Large-Model Reasoning and Real-Time Control via Agentic Fast-Slow Planning}

\author{~~~Jiayi Chen$^{1}$, Shuai Wang$^{2,*}$, Guangxu Zhu$^{1,*}$, and Chengzhong Xu$^{3}$
\thanks{This work was supported in part by National Natural Science Foundation of China (Grant No. 62522118, 62371313),  in part by Shenzhen-Hong Kong-Macau Technology Research Programme (Type C) (Grant No. SGDX20230821091559018), in part by the Shenzhen Science and Technology Program (Grant No. JCYJ20241202124934046),  in part by Guangdong Young Talent Research Project (Grant No. 2023TQ07A708). \protect\\
\makebox[1.5em]{}$^{*}$Corresponding authors: Shuai Wang and Guangxu Zhu.\protect\\
\makebox[1.5em]{}$^{1}$Jiayi Chen and Guangxu Zhu are with the Shenzhen Research Institute of Big Data, The Chinese University of Hong Kong (Shenzhen), Shenzhen 518115, China ({\tt\footnotesize jiayichen5@link.cuhk.edu.cn, gxzhu@sribd.cn}). 
$^{2}$Shuai Wang is with the Shenzhen Institutes of Advanced Technology (SIAT), Chinese Academy of Sciences, Shenzhen, China ({\tt\footnotesize s.wang@siat.ac.cn}). 
$^{3}$Chengzhong Xu is with the State Key Laboratory of Internet of Things for Smart City (SKL-IOTSC), University of Macau, Macau, China.}
}

\begin{document}

\maketitle
\thispagestyle{empty}
\pagestyle{empty}

\begin{abstract}
Large foundation models enable powerful reasoning for autonomous systems, but mapping semantic intent to reliable real-time control remains challenging. Existing approaches either (i) let Large Language Models (LLMs) generate trajectories directly—brittle, hard to verify, and latency-prone—or (ii) adjust Model Predictive Control (MPC) objectives online—mixing slow deliberation with fast control and blurring interfaces. We propose \textbf{Agentic Fast–Slow Planning}, a hierarchical framework that decouples perception, reasoning, planning, and control across natural timescales. The framework contains two bridges. \textbf{Perception2Decision} compresses scenes into ego-centric topologies using an on-vehicle Vision–Language Model (VLM) detector, then maps them to symbolic driving directives in the cloud with an LLM decision maker—reducing bandwidth and delay while preserving interpretability.  \textbf{Decision2Trajectory} converts directives into executable paths: Semantic-Guided A$^{*}$ embeds language-derived soft costs into classical search to bias solutions toward feasible trajectories, while an Agentic Refinement Module adapts planner hyperparameters using feedback and memory. Finally, MPC tracks the trajectories in real time, with optional cloud-guided references for difficult cases. Experiments in CARLA show that Agentic Fast–Slow Planning improves robustness under perturbations, reducing lateral deviation by up to 45\% and completion time by over 12\% compared to pure MPC and an A$^{*}$-guided MPC baseline. Code is available at \url{https://github.com/cjychenjiayi/icra2026_AFSP}.

\end{abstract}


\section{Introduction}

Large foundation models are increasingly used as reasoning engines in autonomous systems, offering broad knowledge and flexible problem-solving \cite{tian2024_drivevlm, wen2024_dilu}. The core challenge is translating high-level semantic intent into reliable real-time control. Existing approaches mainly follow two routes, each with limitations. One lets Large Language Models (LLMs) generate trajectories directly \cite{yang2025_trajllm}. While flexible, such outputs are brittle, hard to verify, and difficult to execute within strict timing constraints. The other adapts Model Predictive Control (MPC) objectives or parameters via language \cite{sha2023_languagempc}, but this mixes slow deliberation with fast control, blurring the interface between reasoning, planning, and actuation.

These limitations reveal a deeper gap: current strategies blur the distinct roles and timescales of perception, reasoning, planning, and control instead of exploiting their complementary strengths \cite{saha2025_system1x}. A key design principle in our work is \textit{scale-aware decoupling} \cite{ding2021_epsilon}: decomposing the pipeline so that each layer operates at its appropriate rate with explicit, interpretable interfaces.  

\begin{figure}[tp]
\centering
\includegraphics[width=0.95\linewidth]{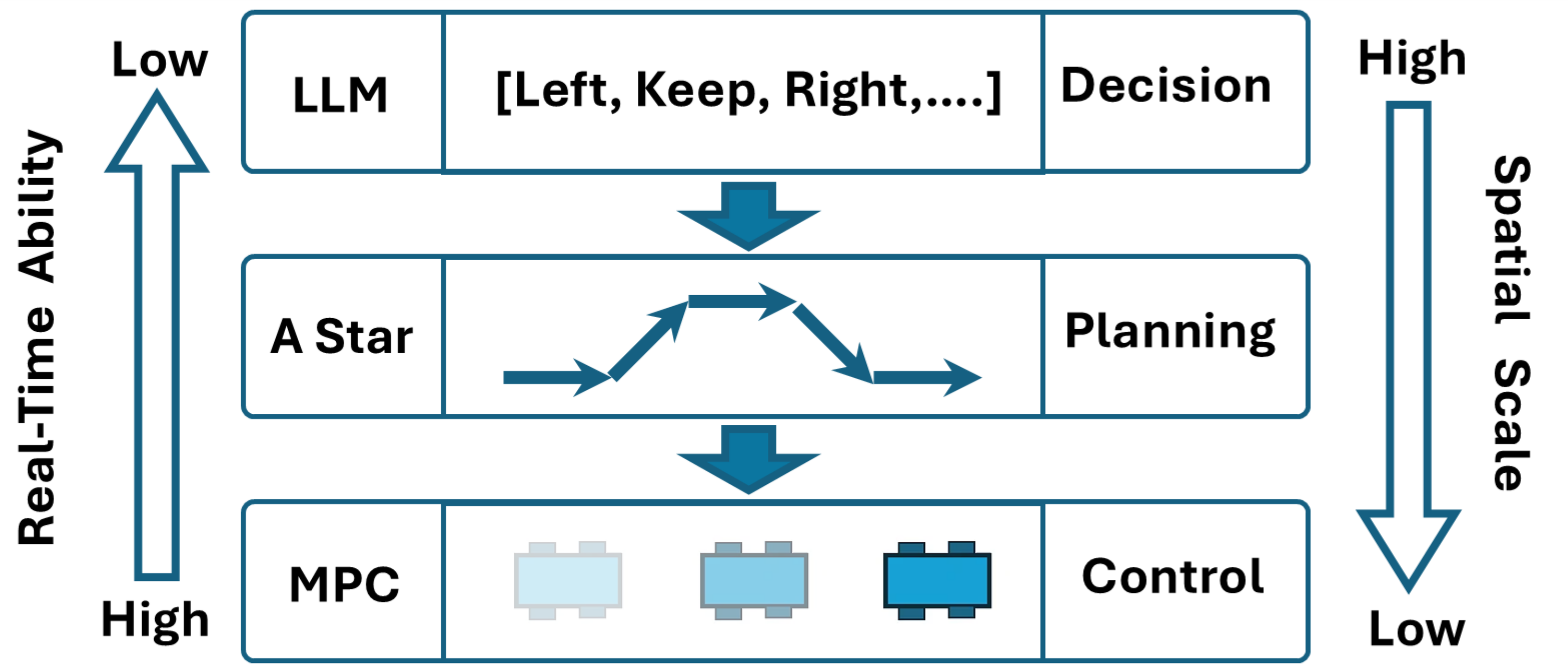}
\vspace{-0.05in}
\caption{Overview of Agentic Fast--Slow Planning Hierarchy}
\label{Fig:collaboration}
\vspace{-0.1in}
\end{figure}

We propose \textbf{Agentic Fast--Slow Planning (AFSP)}, a hierarchical framework inspired by dual-process cognition that separates slow reasoning, mid-level planning, and fast closed-loop control. As shown in Fig.~\ref{Fig:collaboration}, the three layers are: (i) a reasoning layer that interprets scene and issues symbolic decisions, (ii) a planner that converts these into trajectories \cite{karaman2011_rrtstar}, and (iii) a control layer that ensures feasibility and safety in real time \cite{yu2021_mpc_review}. This division lets language models focus on semantic reasoning, planners on trajectory generation, and MPC on stable execution. While the interface from A$^{*}$ to MPC is well studied \cite{yu2021_mpc_review}, the bridge from LLM outputs to planners remains underexplored \cite{yang2025_trajllm, sha2023_languagempc}.

We identify two challenges at the LLM-to-planner interface and address them with two modules. First, edge resource limits make transmitting or processing raw perceptual inputs inefficient. Our \textbf{Perception2Decision} module combines an on-vehicle Vision--Language Model (VLM) Topology Detector with a cloud-based LLM Decision Maker \cite{wen2024_dilu}. VLM encodes images into compact ego-centric topology graphs, while the LLM uses chain-of-thought and in-context prompting to output symbolic directives. This preserves low-latency perception on vehicle, shifts heavy reasoning to the cloud, providing interpretable intent with minimal bandwidth. 

Second, classical A$^{*}$ is sensitive to hyperparameters and map variations, making it brittle under perturbations \cite{hart1968_astar}. Our \textbf{Decision2Trajectory} module addresses this with two components. \textit{Semantic-Guided A$^{*}$} adds language-derived soft costs that bias paths toward intent-consistent, feasible solutions without brittle constraints. An \textit{Agentic Refinement Module} adaptively tunes planner hyperparameters with feedback and memory, automating the trial-and-error usually requiring human expertise. Together, these preserve classical search’s geometric feasibility and verifiability while improving robustness and adaptability.

In sum, our contributions are threefold: (i) a hierarchical framework—\textbf{Agentic Fast--Slow Planning}—that bridges reasoning, planning, and control across timescales; (ii) Perception2Decision, which links perception to symbolic decision-making under edge--cloud constraints; and (iii) Decision2Trajectory, which combines semantic guidance with agentic refinement to produce robust, intent-consistent trajectories. This design keeps slow deliberation out of the real-time loop, provides interpretable intermediates, and leverages the strengths of learning and optimization. Experiments in CARLA show robustness under perturbations, reducing lateral deviation by up to 45\% and completion time by over 12\% compared to MPC and an A$^{*}$-guided MPC baseline.

\begin{figure*}[t]
\centering
\includegraphics[width=0.97\linewidth]{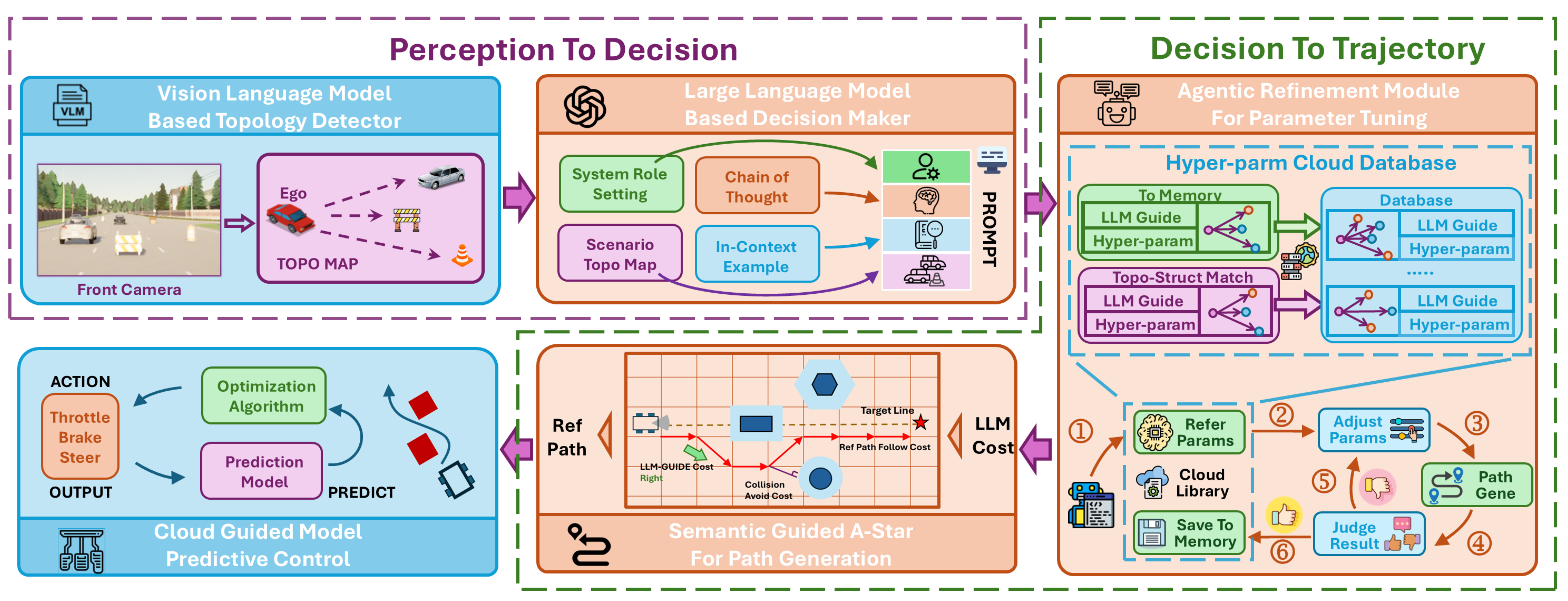}
\vspace{-0.07in}
\caption{System architecture of Agentic Fast--Slow Planning}
\label{Fig.structure}
\vspace{-0.22in}
\end{figure*}

\section{Related Works}
\subsection{Optimization-Based Autonomous Driving System}
Classical autonomous driving stacks rely on search and optimization, where explicit dynamics and constraints ensure deterministic behavior and real-time guarantees. Graph-based methods like A$^{*}$ \cite{hart1968_astar} and RRT$^{*}$ \cite{karaman2011_rrtstar} support discrete-space planning, while optimization-based approaches such as MPC \cite{yu2021_mpc_review}, RDA \cite{han2023_rda}, and spatio-temporal corridor methods like EPSILON \cite{ding2021_epsilon} handle constrained trajectory optimization. Despite these advances, pipelines depend on hand-crafted costs and extensive hyperparameter tuning, limiting adaptability and robustness in long-tail scenarios.

\subsection{Learning-Based Autonomous Driving System}
Learning-based approaches unify perception, reasoning, and planning in a single pipeline, often leveraging VLMs or LLMs with external knowledge. Multi-modal methods such as DriveVLM \cite{tian2024_drivevlm} couple perception with spatial reasoning, while Trajectory-LLM \cite{yang2025_trajllm} uses language-guided trajectory generation to expand data coverage. DiLu \cite{wen2024_dilu} adds rule- and memory-based mechanisms for long-tail robustness, and KoMA \cite{jiang2024_koma} adopts multi-agent LLMs with shared memory for decision-making. Knowledge-grounded designs also emerge: ODA \cite{sun2024_oda} injects knowledge graphs into planning, optical-flow distillation links semantics to control \cite{wang2021_ivmp} and surveys highlight both
the potential and scalability concerns of graph-driven planning \cite{luettin2022_kg_ad_survey} Despite their flexibility, learning-based ADS struggle with control rates, and their opaque reasoning-to-actuation pathways limit interpretability and safety assurance.

\subsection{Hybrid-Based Autonomous Driving System}
Hybrid approaches couple learning-based intent with classical planners and controllers to balance efficiency and safety. LanguageMPC \cite{sha2023_languagempc} maps language commands to MPC objectives, while MPCxLLM \cite{baumann2025_mpcxllm} adapts on-board MPC with task-level input from an LLM. System~1.x \cite{saha2025_system1x} gates costly deliberation, invoking slow reasoning selectively while a fast controller closes the loop, and DriveVLM \cite{tian2024_drivevlm} likewise exploits a slow--fast split for scheduling. LVM-MPC \cite{chen2025_ocp} coordinates when to query a remote LVM and how to fuse its guidance with local MPC. Yet most hybrids reduce LLMs to one-shot hyperparameter updates, lacking principled interfaces to control objectives or query timing, thus underutilizing learning–optimization complementarity.

\section{System Overview}
The input to Agentic Fast--Slow Planning consists of raw sensory observations (front camera images) together with navigation goals, while the outputs are real-time control actions (i.e., throttle and steering). As shown in Fig.~\ref{Fig.structure}, the goal of Agentic Fast--Slow Planning is to map high-level reasoning into reliable execution by bridging perception, planning, and control across distinct timescales, thereby improving trajectory robustness and interpretability while preserving real-time performance.

\textbf{Perception2Decision} spans edge and cloud. On-vehicle, a lightweight \textbf{VLM-based Topology Detector} encodes sensory inputs into an ego-centric topology $\mathcal{Z}$, reducing data volume while preserving semantics. In the cloud, an \textbf{LLM-based Decision Maker} consumes $\mathcal{Z}$ to produce a decision sequence $S$, offloading high-level reasoning yet keeping perception local. These decisions provide interpretable intent but must be further translated into executable trajectories.

\textbf{Decision2Planning} translates high-level LLM guidance into planning paths, leveraging a robust yet efficient approach.
\textbf{Semantic-Guided A$^{*}$} translates symbolic decisions into reference trajectories $\tau^\star$ by embedding language-derived costs into classical search, ensuring trajectories remain feasible and semantically consistent. To enhance adaptability, an \textbf{Agentic Refinement Module} tunes the planner’s hyperparameters online using structured feedback and past experience, allowing the system to adjust to scene variability without manual retuning. Together, these components couple symbolic reasoning with heuristic planning in a robust and interpretable manner.
The finalized reference trajectory $\tau^\star$ is then tracked on the vehicle by \textbf{Cloud-Guided MPC} to guarantee real-time feasibility and responsiveness.

\section{Perception to Decision}

Directly feeding raw images to large VLMs is inefficient: high-dimensional inputs require heavy visual backbones (e.g., ViTs) and incur unnecessary bandwidth, although driving decisions depend on spatial relations rather than fine-grained appearance. To avoid this overhead, \textbf{Perception2Decision} separates perception and reasoning. A lightweight on-vehicle \textbf{VLM-based Topology Detector} extracts an ego-polar topology encoding object semantics and spatial geometry, while a cloud \textbf{LLM-based Decision Maker} reasons over the compact graph to generate symbolic driving directives. This decoupling preserves low-latency edge perception, leverages cloud reasoning, and provides a structured interface for downstream planning and control.

\subsection{VLM-based Topology Detector}

We aim to compress each scene into a topology graph that preserves spatial and semantic structure while discarding redundant visual detail, so that downstream reasoning can operate directly on graphs rather than raw images. This design reduces bandwidth, improves efficiency, and provides an interpretable interface for decision-making. Concretely, a scene is encoded as $G=\{z_j\}$, where each $z_j=(b_j,d_j,\phi_j,t_j)$ contains bounding box $b_j$, ego-polar distance $d_j$, orientation $\phi_j$, and semantic class $t_j\in\mathcal{C}$. Ego-polar coordinates are obtained from world-frame positions $(X_j,Y_j)$ by rotation into the ego frame with yaw $\theta_{\mathrm{ego}}$:
\begin{equation}
(d_j,\phi_j)=R(-\theta_{\mathrm{ego}})\,(X_j-X_{\mathrm{ego}},\,Y_j-Y_{\mathrm{ego}}).
\end{equation}
To enhance robustness, distances are quantized to $0.1$ and orientations to $0.5^\circ$, with $0^\circ$ along the forward axis.

\begin{figure}[tp]
\centering
\includegraphics[width=0.95\linewidth]{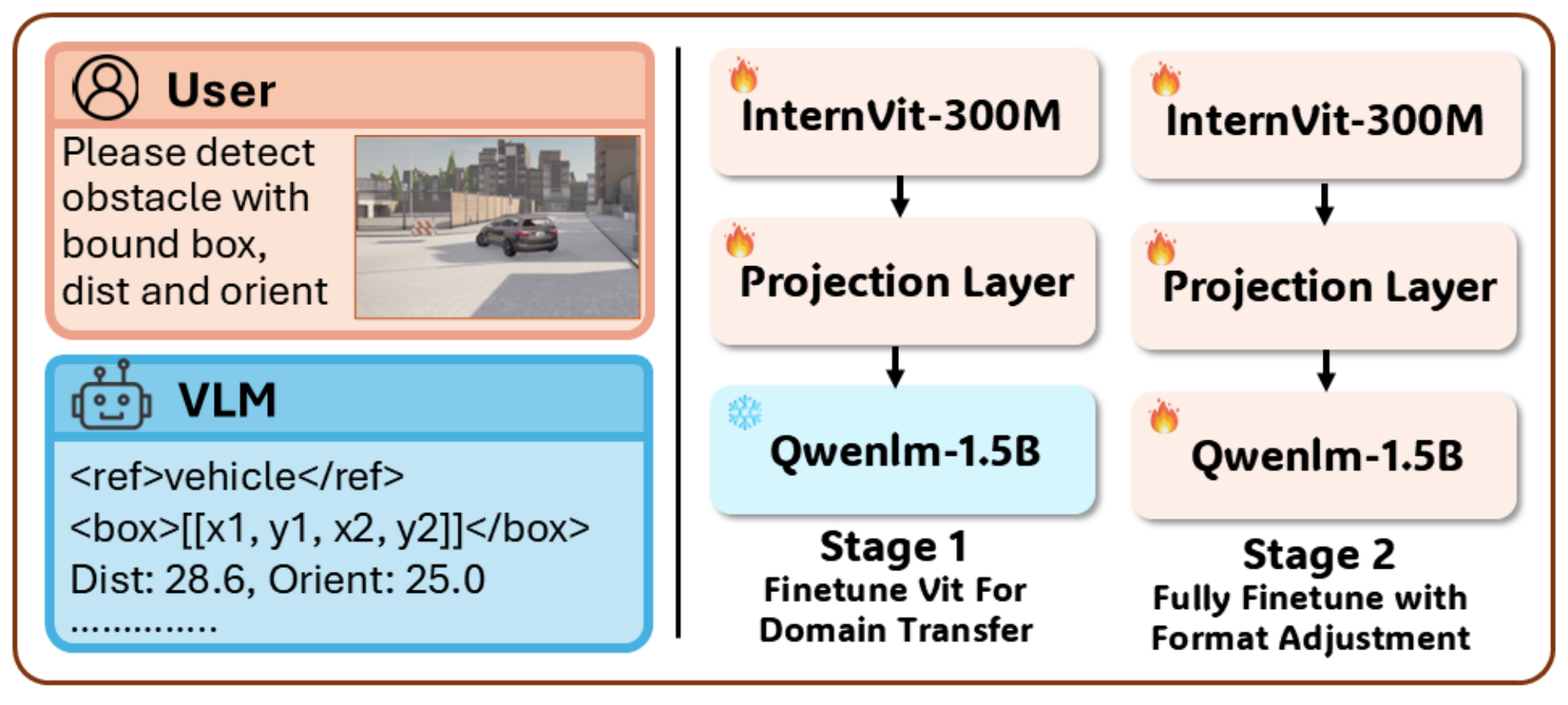}
\vspace{-0.1in}
\caption{VLM Fine-tuning data format and strategy.}
\label{fig:finetune_strategy}
\vspace{-0.1in}
\end{figure}

During training, the topology graph $G$ is serialized into a structured text sequence where each object is described by semantic class, bounding box, and spatial attributes. Special tokens (\texttt{<ref>}, \texttt{<box>})~\cite{liu2023_llava} enforce consistent localization, while distance and orientation are expressed in plain text. We fine-tune a VLM with 2B parameters using a two-stage strategy (Fig.~\ref{fig:finetune_strategy}): Stage~1 freezes the language backbone to adapt the vision encoder and projection layers for stable grounding; Stage~2 unfreezes all parameters for joint alignment. Given image $I$ and target sequence $Y=(y_1,\ldots,y_T)$, the model is trained with next-token prediction:
\begin{equation}
\mathcal{L}(\theta)=-\sum_{t=1}^{T}\log p_{\theta}(y_t \mid I, y_{<t}),
\end{equation}
where $\theta$ denotes trainable parameters in the current stage. This progressive adaptation preserves pretrained language priors, activates spatial reasoning, and produces interpretable topology graphs for downstream planning.

\subsection{LLM-Based Decision Maker}

Directly producing low-level trajectories or controller parameters from an LLM is impractical due to precision, safety, and timing constraints. Instead, we restrict its role to \emph{semantic-level decision-making}, providing an interpretable interface between reasoning and planning. The LLM consumes the serialized topology graph from the \textbf{VLM-based Topology Detector} and outputs a compact driving plan: a short sequence of discrete directives (e.g., \textsc{left}, \textsc{right}, \textsc{keep}) with an associated driving style mapped to reference velocity. Since any maneuver can be decomposed into such primitives, the representation remains expressive.

To ensure reliability, we adopt structured prompting with three elements: a role description fixing the LLM as a driving planner, in-context examples constraining the format, and a chain-of-thought process that enumerates obstacles, assigns risk levels, and derives safe directives. As shown in Fig.~\ref{Fig:result_lm}, this design produces symbolic decisions capturing the semantic structure of driving scenes while suppressing hallucinations. Outputs follow a fixed schema with fields such as \texttt{Reasoning}, \texttt{Drive Plan}, and \texttt{Drive Style}, enabling direct parsing by the downstream \textbf{Semantic-Guided A$^{*}$} planner. Compared with direct image-based VLM inference, this graph-to-decision pathway achieves similar decision quality with lower latency, removing redundant visual detail while preserving essential spatial relations.

\section{Decision to Trajectory}

Classical search-based planners such as A$^{*}$ remain attractive for autonomous driving due to their geometric feasibility and verifiability, yet they are rarely used in practice because of sensitivity to hyperparameters, shortsighted search, and susceptibility to local optima. More sophisticated approaches, such as spatio-temporal corridors, introduce high complexity, while learning-based planners often lack interpretability and stability under real-time constraints.

To address these limitations, we retain the \textbf{Decision2Trajectory} framework with two key modules. First, \textbf{Semantic-Guided A$^{*}$} incorporates language-derived costs to provide strong semantic guidance, improving robustness and consistency without altering the underlying search structure. Second, an \textbf{Agentic Refinement Module} mitigates parameter sensitivity by maintaining memory of past settings and performing online self-adjustment. Together, these components preserve the strengths of classical search while enhancing adaptability and stability in dynamic driving scenarios.

\begin{figure*}[t]
  \centering
  \begin{minipage}[t]{0.27\textwidth}
    \centering
    \includegraphics[height=5cm]{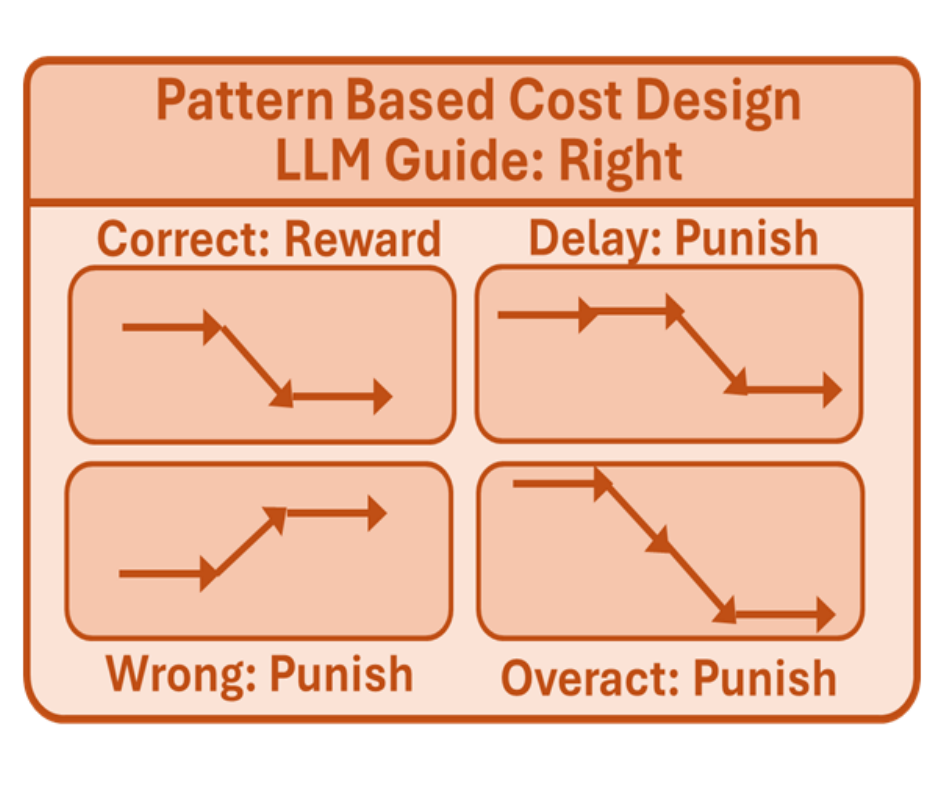}
    \vspace{-0.3in}
    \caption*{(a) Pattern cost design}
  \end{minipage}%
  \hfill
  \begin{minipage}[t]{0.68\textwidth}
    \centering
    \includegraphics[height=4.8cm]{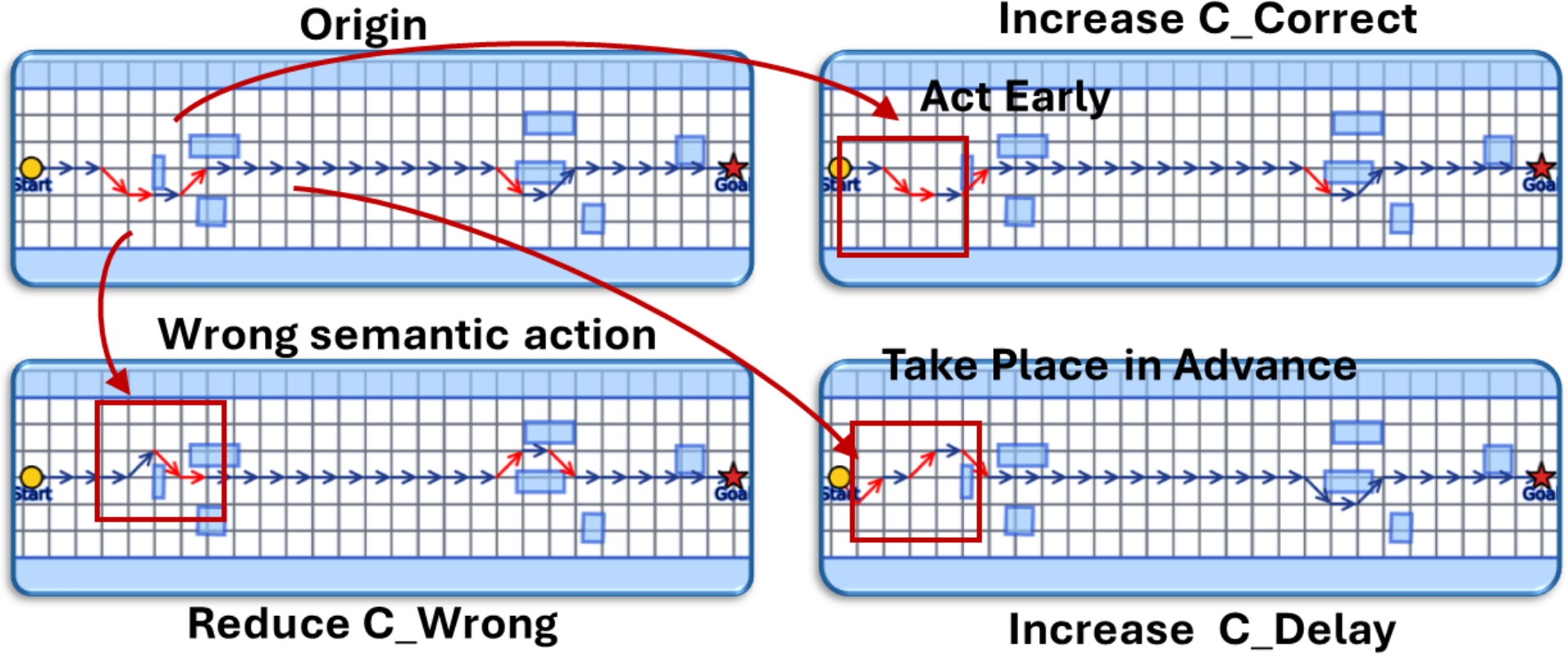}
    \vspace{-0.1in}
    \caption*{(b) Observation of hyperparameter influence}
  \end{minipage}
  \caption{Illustration of Semantic-Guided A*: (a) Pattern cost design and (b) observation of hyperparameter influence}
  \label{Fig:observation}
  \vspace{-0.15in}
\end{figure*}

\subsection{Semantic-Guided A\texorpdfstring{$^{*}$}{*} Path Planning}
\label{sec:semanticastar}
In classical A$^{*}$ search, each node is defined by grid coordinates $(i,j)$ and evaluated as $f(s)=g(s)+h(s)$, where $g$ is the accumulated geometric cost and $h$ a heuristic estimate. While effective for feasibility, such paths are agnostic to high-level intent and may miss symbolic directives, e.g., delayed lane changes or missed turns.

We propose a \emph{Semantic-Guided A$^{*}$} algorithm that augments the classical formulation with \emph{soft semantic costs}. Rather than enforcing LLM-derived directives as hard constraints (risking dead ends), we encode them as additional costs. This biases search toward intent-consistent trajectories while preserving robustness and allowing recovery when directives conflict with geometry or safety.

Concretely, we extend the state representation with two variables: previous move $m_{\text{prev}} \in \{\text{F},\text{FL},\text{FR}\}$ and an index $k$ tracking progress through the LLM-supplied directive sequence $\mathcal{A} = (a_1,\dots,a_T)$, where $a_t \in \{\text{left},\text{right},\text{keep}\}$. At each expansion, the transition $(m_{\text{prev}}, m_{\text{curr}}, a_k)$ \cite{ding2021_epsilon} is mapped by translation logic $\Phi$ into four alignment categories—\emph{Correct} (realization of $a_k$), \emph{Delay} (continuing without realizing $a_k$), \emph{Wrong} (contradicting $a_k$), or \emph{Overact} (repeating $a_k$). Categories are assigned semantic step costs:
\begin{equation}
c_{\text{sem}} \in \{C_{\text{CORR}},\,C_{\text{DELAY}},\,C_{\text{WRONG}},\,C_{\text{OVER}}\},
\end{equation}
where $C_{\text{CORR}} < 0$ rewards alignment, $C_{\text{WRONG}} \gg 0$ penalizes contradictions, and $C_{\text{DELAY}}, C_{\text{OVER}}$ impose mild penalties. This soft-cost design biases search toward intent-consistent paths while still allowing recovery from directive errors or local conflicts. The accumulated cost update is
\begin{equation}
g(s') = g(s) + c_{\text{geom}}(s,s') + c_{\text{sem}}(m_{\text{prev}}, m_{\text{curr}}, a_k),
\end{equation}
while the heuristic $h$ retains the classical geometric potential-field terms. Once a directive $a_k$ is realized, the index advances as $k \leftarrow k+1$. A valid solution requires $k=T$ at the goal, ensuring that all semantic instructions are completed.

This formulation extends the classical $f(s)=g(s)+h(s)$ by embedding LLM-derived directives into the accumulated cost, while keeping the state compact with polynomial complexity. The planner outputs both feasible paths and interpretable markers of directive realization, supporting downstream tuning. To the best of our knowledge, this is the first principled extension of A$^{*}$ that incorporates language-derived semantics as soft costs, enabling intent-consistent yet robust path planning and establishing a novel interface between symbolic reasoning and graph search.

\begin{algorithm}[!t]
\caption{Semantic-guided A$^{*}$ Path Generation}
\KwIn{Current state $(i,j,m_{\text{prev}},k)$; directive sequence $\mathcal{A}$; hyperparameters $(C_{\text{CORR}},C_{\text{DELAY}},C_{\text{WRONG}},C_{\text{OVER}})$}
\ForEach{neighbor $(i',j')$ of $(i,j)$}{
    $m_{\text{curr}} \leftarrow \text{MoveLabel}((i,j),(i',j'))$\;
    $a \leftarrow \mathcal{A}[k]$ \tcp*{current directive}
    $(c_{\text{sem}}, \text{completed}) \leftarrow \text{SemanticStepCost}(m_{\text{prev}}, (i,j),(i',j'), a)$\;
    $c_{\text{step}} \leftarrow c_{\text{geom}}(s,s') + c_{\text{sem}}$\;
    $g(i',j') \leftarrow g(i,j) + \max\{c_{\text{step}}, \epsilon\}$ \tcp*{ensure non-negative increment}
    $k' \leftarrow k+1$ if completed else $k$\;
    Push state $(i',j',m_{\text{curr}},k')$ into open set with score $f=g+h$\;
}
\end{algorithm}

\subsection{Agentic Refinement Module For Parameter Tuning}

Classical planners such as A$^{*}$ inherently rely on a set of cost weights or hyperparameters. 
From Fig.~\ref{Fig:observation} we observe that these hyperparameters exert distinct influences on trajectory shape: 
increasing $C_{\text{CORR}}$ encourages aggressive realization of directives, 
higher $C_{\text{DELAY}}$ tolerates postponed actions at the risk of missing turns, 
$C_{\text{WRONG}}$ enforces semantic consistency but often causes detours, 
while $C_{\text{OVER}}$ regulates oscillatory behavior from repeated actions. 
No single setting works across all scenarios, and manual tuning is infeasible. 
This motivates \textbf{Agentic Refinement Module}, a closed-loop process where human adjustment logic—perturb parameters, 
observe feedback, and refine—can be delegated to an agent.

We design a core prompt that encodes this human-like tuning loop: it incorporates warm-start parameters, semantic guidance $g$, and planner feedback, and instructs the LLM to refine hyperparameters $\theta$ and determine whether further retries are required. In essence, human trial-and-error is operationalized into structured prompt reasoning.

The framework leverages three tools. 
\texttt{SelectRefHyperparams} retrieves warm-start parameters by matching current scenes to similar cases in a cloud memory. 
\texttt{AStarPathGenerate} executes the semantic-guided A$^{*}$ planner (Section~\ref{sec:semanticastar}) and reports trajectories with structured metrics. 
\texttt{SaveScene} logs successful $(s,g,\theta^*)$ triplets, enabling continual knowledge accumulation across scenarios. This design mirrors our implementation: 
\texttt{SelectRefHyperparams} provides knowledge-driven warm starts, 
\texttt{AStarPathGenerate} evaluates hyperparameters through semantic-guided planning, 
and \texttt{SaveScene} stores refined configurations. 
The LLM prompt coordinates these tools in a closed loop, automating parameter refinement that would otherwise require manual tuning. 
This agentic process improves adaptability across diverse scenarios and enables continual knowledge accumulation without human intervention.

\begin{algorithm}[!t]
\caption{Agentic Refinement Module}
\KwIn{Scene $s$, semantic guidance $g$, max retries $k$}
$\theta_0 \leftarrow$ \texttt{SelectRefHyperparams($s$)} \tcp*{warm-start}
\For{$i=1$ \KwTo $k$}{
    $(traj,metrics) \leftarrow$ \texttt{AStarPathGenerate($g,\theta_{i-1},i$)}\;
    \If{Acceptable(metrics)}{
        \texttt{SaveScene($s,g,\theta_{i-1}$)}\;
        \textbf{break}\;
    }
    $\theta_i \leftarrow$ LLM\_Refine($\theta_{i-1}, metrics$)\;
}
\end{algorithm}

\subsection{Cloud-Guided Model Predictive Control}

To bridge the reference path generated by \textbf{Decision2Planning}, we design a \textbf{Cloud-Guided MPC} in which:
$\mathcal{H}=\{t,\ldots,t+H\}$. At each step, the controller selects either a local lightweight reference $\tau^{\text{local}}$ or a cloud-provided global reference $\tau^{\text{cloud}}$, using a binary indicator $z_t \in \{0,1\}$. The predicted trajectory must satisfy dynamics and safety:
\begin{equation}
s_{t+h}^\star = (1-z_t)\,\tau^{\text{local}}_{t+h} + z_t\,\tau^{\text{cloud}}_{t+h}, 
\quad h\in\mathcal{H}.
\end{equation}
\begin{equation}
s_{t+1} = A_t s_t + B_t u_t + c_t,\;
\operatorname{dist}\big(\mathcal{G}(s_t),\mathcal{O}_m\big)\ge d_0,\;\forall m.
\end{equation}
The stage cost penalizes tracking error and control effort:
\begin{equation}
C=\sum_{h=0}^{H}\|s_{t+h}-s_{t+h}^\star\|^2 
+ \lambda\sum_{h=0}^{H-1}\|u_{t+h}\|^2.
\end{equation}

This switching-based MPC enables low-latency control under nominal conditions ($z_t=0$) while leveraging LLM reasoning in challenging scenarios ($z_t=1$), combining responsiveness with improved global safety.

\begin{figure}[tp]
\centering
\includegraphics[width=0.95\linewidth]{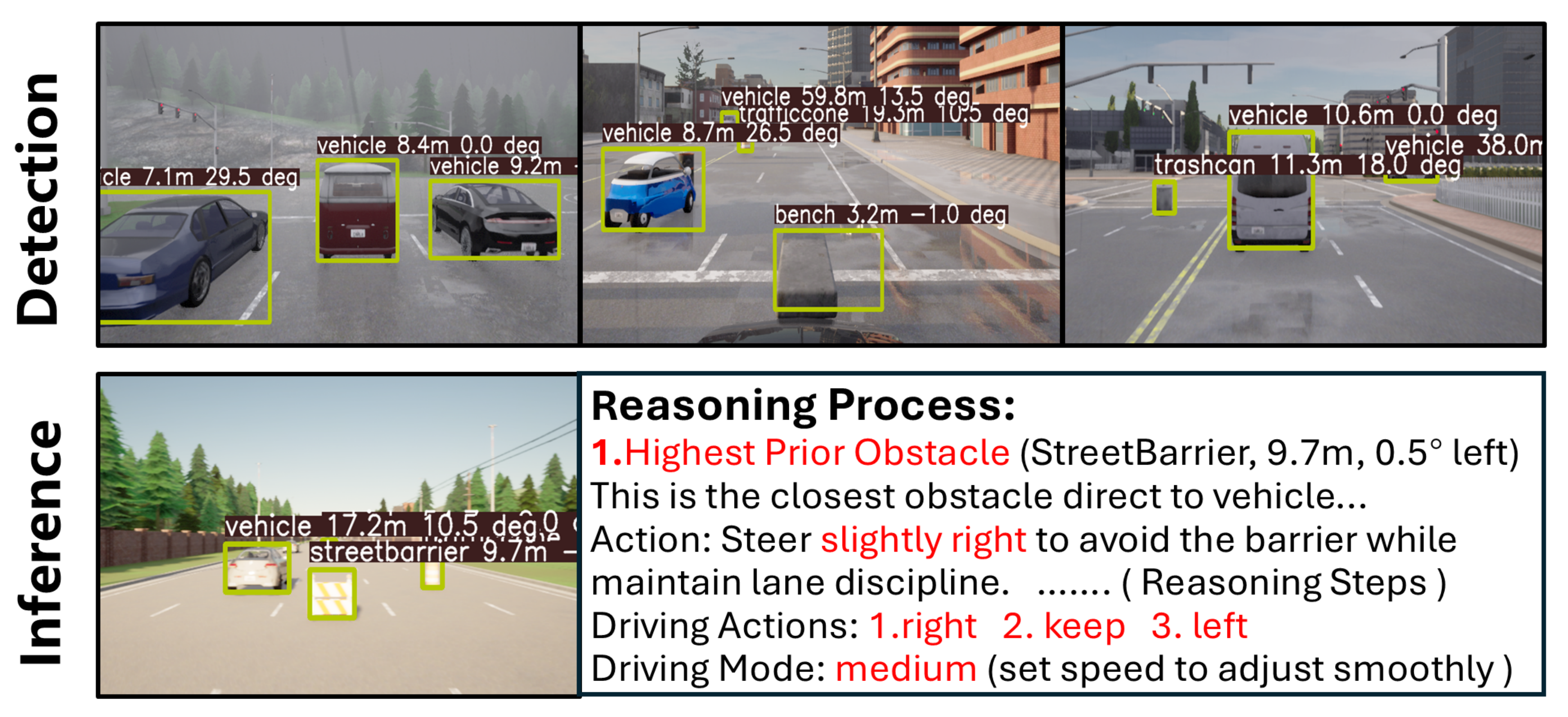}
\vspace{-0.05in}
\caption{Experiment 1: VLM \& LLM output.}
\label{Fig:result_lm}
\vspace{-0.05in}
\end{figure}

\section{Experiments}
\label{experiments}
In this section, we evaluate all the components of the proposed framework of Agentic Fast-Slow Planning.

\subsection{Experiment 1. Evaluation of Perception2Decision}

\subsubsection{Dataset Construction}
We construct a CARLA-based dataset covering \textbf{49 categories}, including vehicles and 48 static or dynamic classes (e.g., barriers, cones, road signs). Frames are collected from Town2--7 and Town10HD, producing $\sim$200{,}000 RGB images. To curate a higher-quality subset, we retain only \textbf{informative frames} where objects are detectable, align bounding boxes with segmentation masks for consistency, and balance sampling across towns. This yields 50{,}000 frames for training and 1{,}000 uniformly sampled frames as the test set. Each frame is annotated with both bounding boxes and a traffic topology graph, where each node stores \textbf{category, distance, orientation, and bounding box}, serialized as question--answer pairs for VLM training.

\begin{table}[h]
\centering
\caption{Performance comparison of fine-tuning strategies. 
Arrows indicate the preferred direction of each metric.}
\label{tab:finetune_new}
\begin{tabular}{lcccc}
\toprule
Method & BBox IoU$\uparrow$ & Cat$\downarrow$ & Dist/Orient$\downarrow$ & Dist Err$\downarrow$ \\
\midrule
Vit Only     & 92.8\% & 0.13\% & 0.42/0.11 & 1.34\% \\
Full Param      & \textbf{93.5\%} & 0.08\% & 0.46/0.11 & 1.41\% \\
Two Stage          & 93.0\% & \textbf{0.04\%} & \textbf{0.41/0.10} & \textbf{1.31\%} \\
\bottomrule
\end{tabular}
\end{table}

We build on InternVL-2B~\cite{chen2024_internvl} and compare three strategies: (i) \emph{Freeze-Language, Fine-tune-Vision}; (ii) \emph{Full-Param}; and (iii) Our \emph{Two-Stage Finetune} (first adapt vision with frozen language, then fine-tune jointly).  Evaluation uses four metrics: \textbf{BBox IoU} (post-matching IoU), \textbf{Cat} (category mismatch rate), \textbf{Dist/Orient} (mean distance and orientation errors), and \textbf{Dist Err} (fraction exceeding a distance-error threshold), jointly reflecting semantic and geometric accuracy. Training is performed on 4$\times$A100 GPUs for 12 epochs with a learning rate of $8\times10^{-5}$, saving checkpoints each epoch and reporting the best. The goal is not to engineer a complex detector but to adapt the foundation model's existing capabilities to simulation-specific distributions.

Table~\ref{tab:finetune_new} shows that \textbf{Our Two-Stage Finetune Method} achieves the best overall balance: highest IoU (93.5\%), lowest category error (0.04\%), and improved distance and orientation accuracy. This confirms that adapting vision first and then unfreezing language yields more precise and robust features, making the VLM suitable for detection. Qualitative examples are shown in Fig.~\ref{Fig:result_lm} (zoom-in used for clarity, showing part of the LLM output).

\begin{figure}[tp]
    \centering
    \subfloat[Score distribution]{\includegraphics[width=0.46\linewidth]{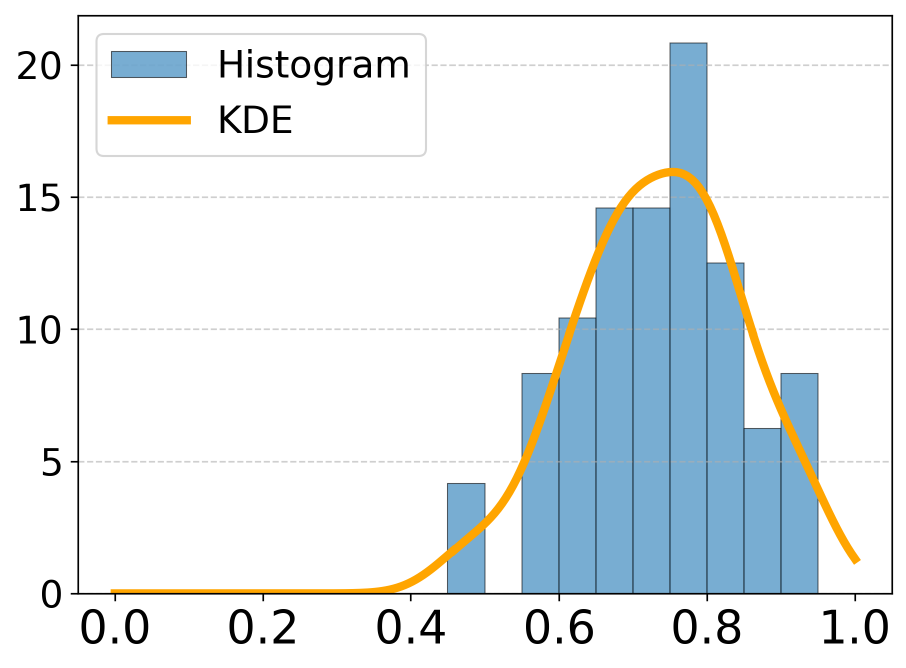}
    \label{match}}
    \subfloat[Time distribution]{\includegraphics[width=0.46\linewidth]{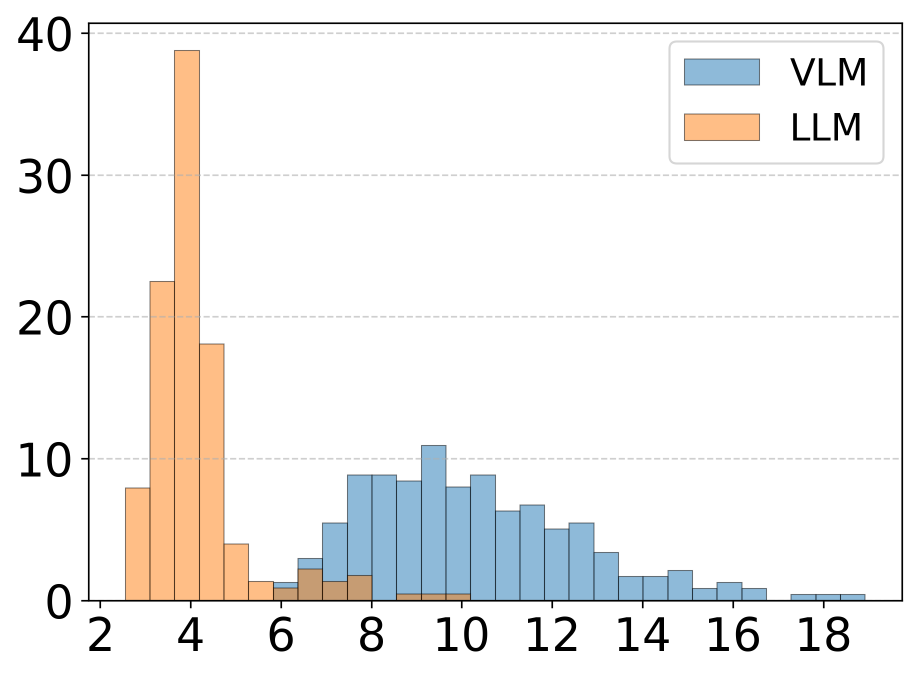}
    \label{time}}
    \caption{Score and time distribution.}
    \label{Fig:compare_vlmllm}
    \vspace{-0.05in}
\end{figure}

\subsubsection{Evaluation of LLM-based Decision Maker}
Symbolic topology graphs are far more efficient to transmit than raw images, making them suitable for cloud--edge pipelines. We evaluate whether a VLM and an LLM give consistent decisions while the LLM achieves faster inference.

For each decision-worthy scene $q$, let
\begin{equation}
\mathcal{A}^{\mathrm{vlm}}_q=\{A^{\mathrm{vlm}}_{q,1},\dots,A^{\mathrm{vlm}}_{q,M}\},
\mathcal{A}^{\mathrm{llm}}_q=\{A^{\mathrm{llm}}_{q,1},\dots,A^{\mathrm{llm}}_{q,N}\},
\end{equation}
be the parsed action sequences. A similarity matrix $S$ is computed and Hungarian assignment solved, yielding
\begin{equation}
\mathrm{Score}(q)=\frac{1}{\min(M,N)}\max_{\pi}\sum_{i=1}^{L_q}\mathrm{sim}(A^{\mathrm{vlm}}_{i},A^{\mathrm{llm}}_{\pi(i)})
\end{equation}

We use \texttt{qwenvl-chat}~\cite{bai2023_qwenvl} (VLM) and \texttt{deepseek-v3}~\cite{deepseek2024_v3} (LLM) via API with same prompt, measuring end-to-end latency. From 200 frames, the top 30\% yielded 48 scenarios. As shown in Fig.~\ref{Fig:compare_vlmllm}, match scores have an average 0.73, while average latency is 4.13\,s for the LLM versus 10.24\,s for the VLM (extreme VLM outliers removed only from histogram). In short, the LLM achieves comparable decision quality with much lower time cost.

\begin{figure}[tp]
\centering
\includegraphics[width=0.95\linewidth]{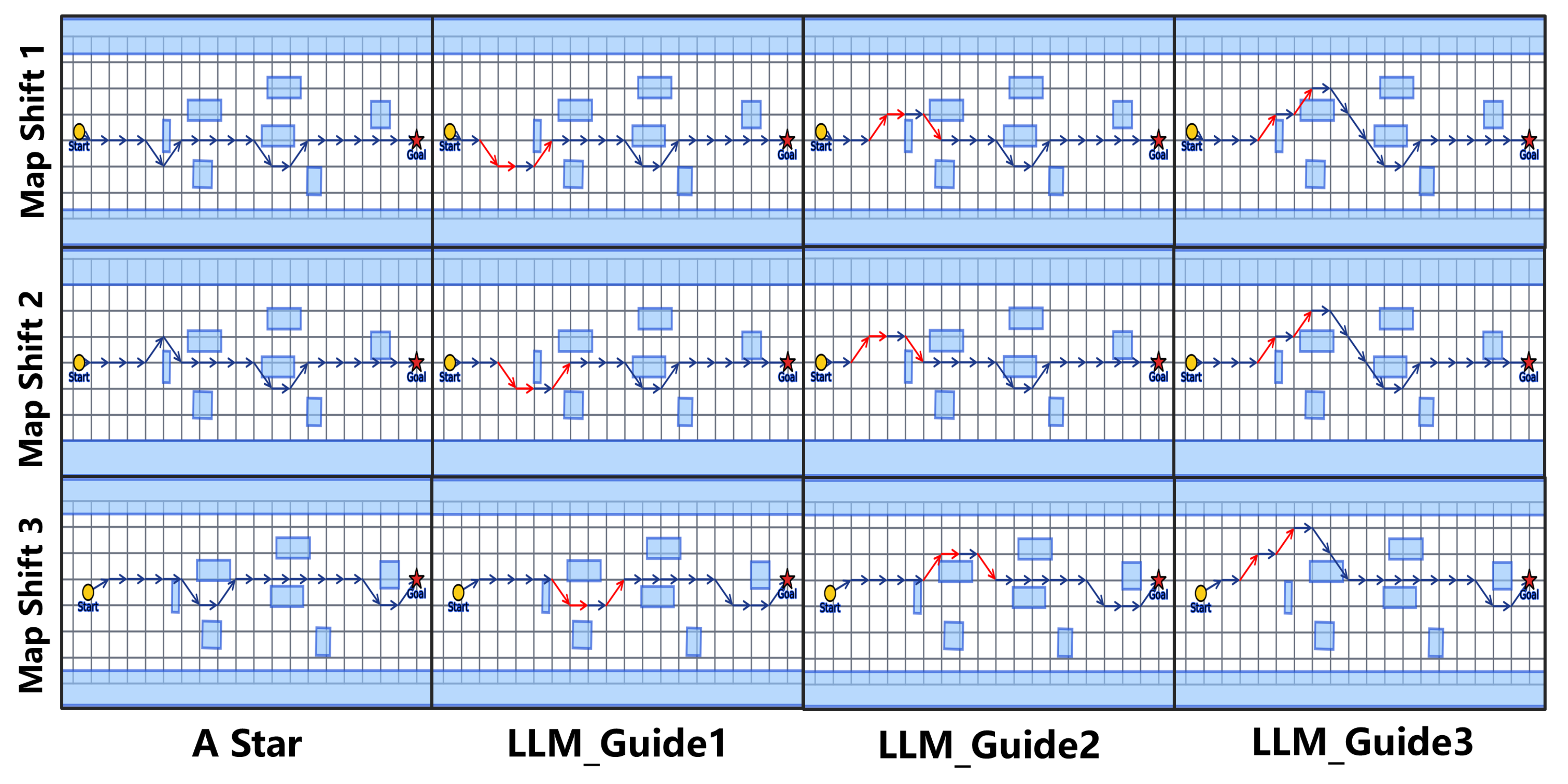}
\caption{Output of semantic-guided A$^{*}$ path generation.}
\label{Fig:result_lmaidastar}
\end{figure}

\subsection{Experiment 2: Evaluation of Decision2Planning}

\subsubsection{Evaluation of Semantic-Guided A\texorpdfstring{$^{*}$}{*}}

We evaluate robustness by testing whether trajectories remain intent-consistent when the grid map is misaligned with the actual start/goal. 
Three settings are considered: 
(i) \textbf{Shift~1}: original map; 
(ii) \textbf{Shift~2}: translated $0.5\,\mathrm{m}$ left; 
(iii) \textbf{Shift~3}: translated $0.5\,\mathrm{m}$ left and $1.0\,\mathrm{m}$ upward. 
\textbf{Start} and \textbf{Goal} shift with the map, while obstacles remain fixed. 
The grid size is $3\,\mathrm{m} \times 3\,\mathrm{m}$.

We compare vanilla A$^{*}$ with three semantic guides: 
G1 = [\textit{right}, \textit{keep}, \textit{left}],
G2 = [\textit{left}, \textit{keep}, \textit{right}], 
G3 = [\textit{left}, \textit{left}], 
keeping all other A$^{*}$ settings unchanged. 
In Shift~1, vanilla A$^{*}$ coincidentally resembles G1, while guided versions follow their intended sequences. 
Under Shift~2 and Shift~3, vanilla A$^{*}$ drifts from the prescribed semantics, whereas guided planners preserve the left/keep/right patterns (red). 
The only exception is G1 under Shift~3 due to accumulated drift; G2 and G3 remain intent-consistent. 
These results demonstrate that semantic guidance enhances robustness to spatial perturbations by preserving high-level intent.

\begin{figure}[tp]
\centering
\includegraphics[width=0.85\linewidth]{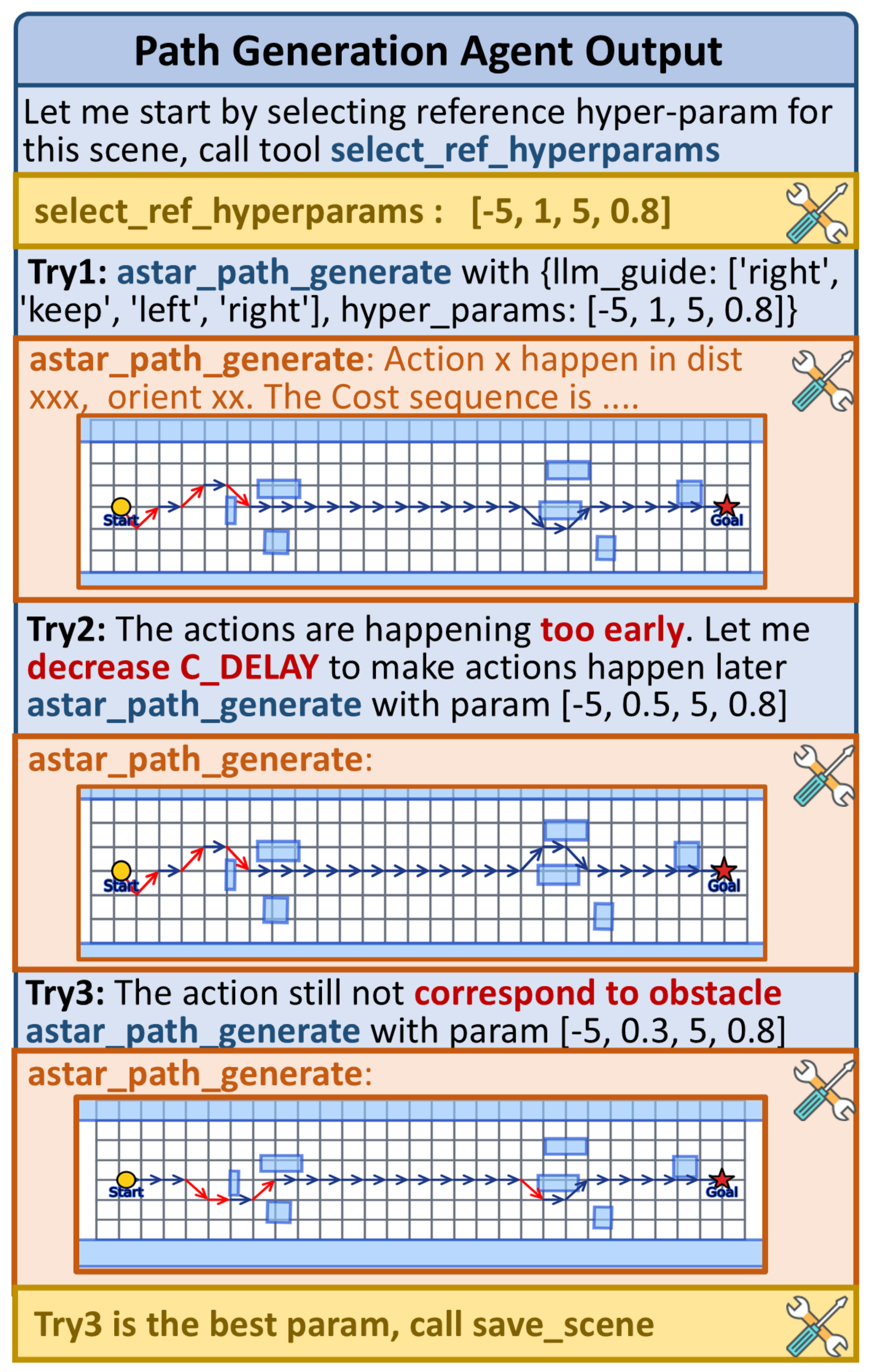}
\vspace{-0.05in}
\caption{Case study of the path generation agent.}
\label{fig:agent_demo}
\vspace{-0.05in}
\end{figure}

\subsubsection{Evaluation of Agentic Refinement Module}

Figure~\ref{fig:agent_demo} shows the closed-loop Agentic Refinement Module. We use deepseek-v3~\cite{deepseek2024_v3} with LangChain to implement the agent flow. Blue boxes denote LLM reasoning, yellow boxes represent cloud memory for parameter storage/retrieval, and orange boxes indicate the trajectory generator.

For stability, the generator does not return full location sequences. Each attempt outputs (i) trial index, (ii) the location triggering semantic cost, and (iii) distance to the obstacle. This grounds semantic actions spatially, while full trajectories are stored separately.

The agent retrieves an initial configuration $[-5,\,1,\,5,\,0.8]$ and generates a trajectory that follows the symbolic sequence but triggers actions prematurely. Reducing $C_{\text{DELAY}}$ to $0.5$ delays actions and improves alignment, though minor mismatch remains. Further refinement to $0.3$ aligns semantic triggers with obstacle boundaries and yields smooth trajectories without oscillation. This configuration is stored as optimal. Overall, the agent integrates reasoning, memory, and trajectory generation to iteratively tune hyperparams. Through semantic feedback and structured penalty analysis, it achieves context-aware refinement without human intervention.

\begin{figure*}[!t]
    \centering
    \subfloat[MPC]{\includegraphics[width=0.31\textwidth]{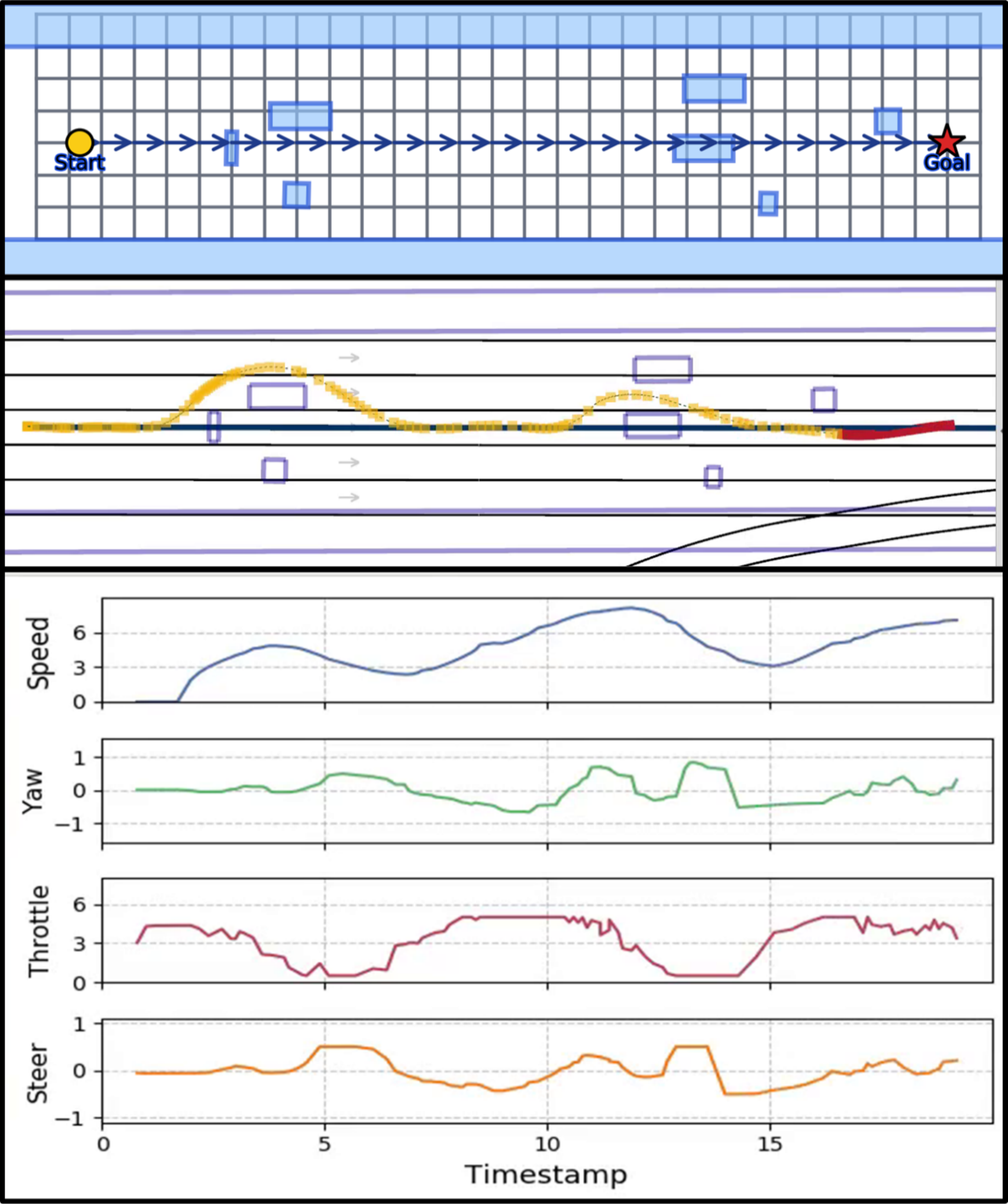}\label{fig:case1_mpc}}
    \hfill
    \subfloat[A$^{*}$-MPC]{\includegraphics[width=0.31\textwidth]{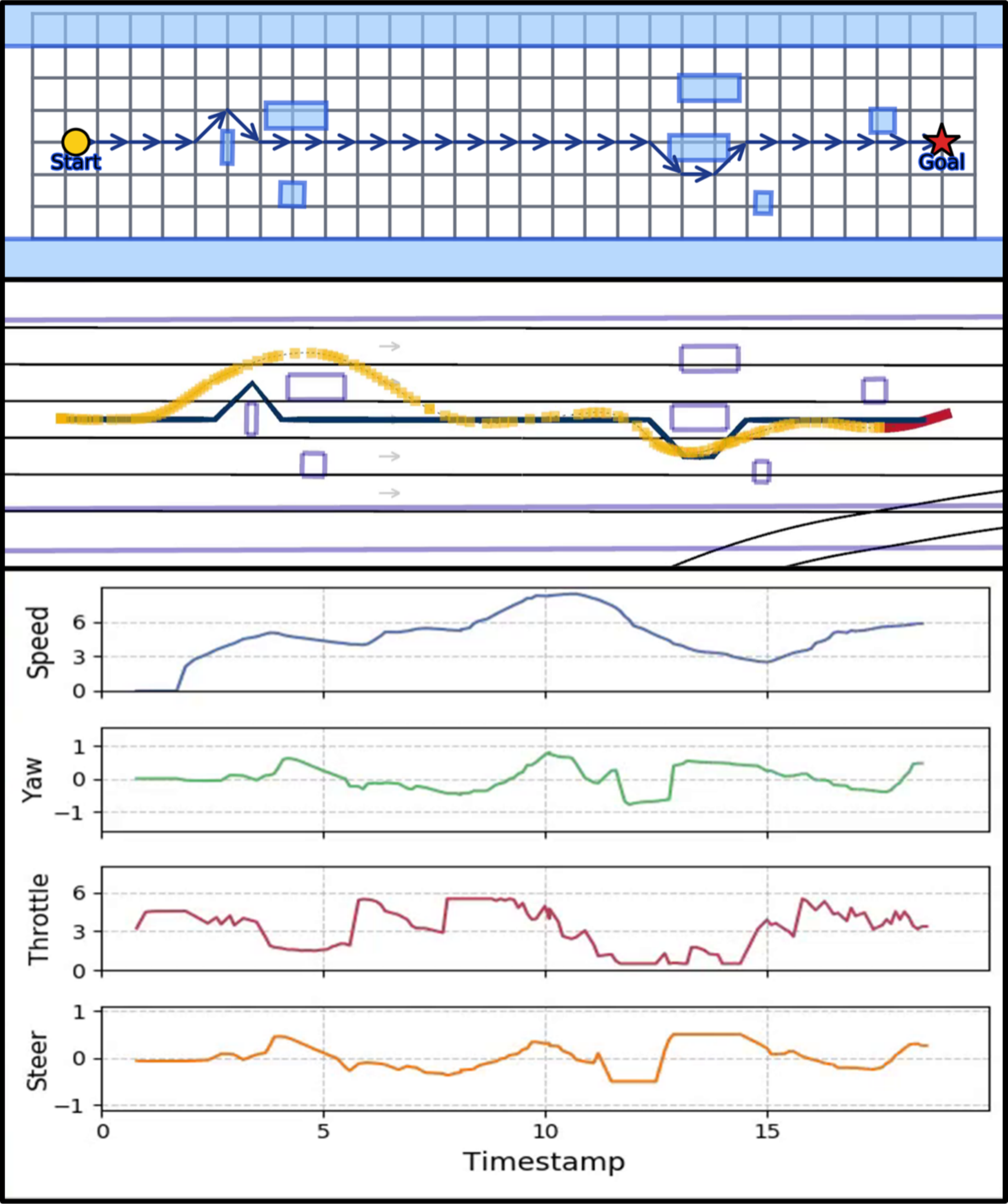}\label{fig:case1_astar}}
    \hfill
    \subfloat[AFSP]{\includegraphics[width=0.31\textwidth]{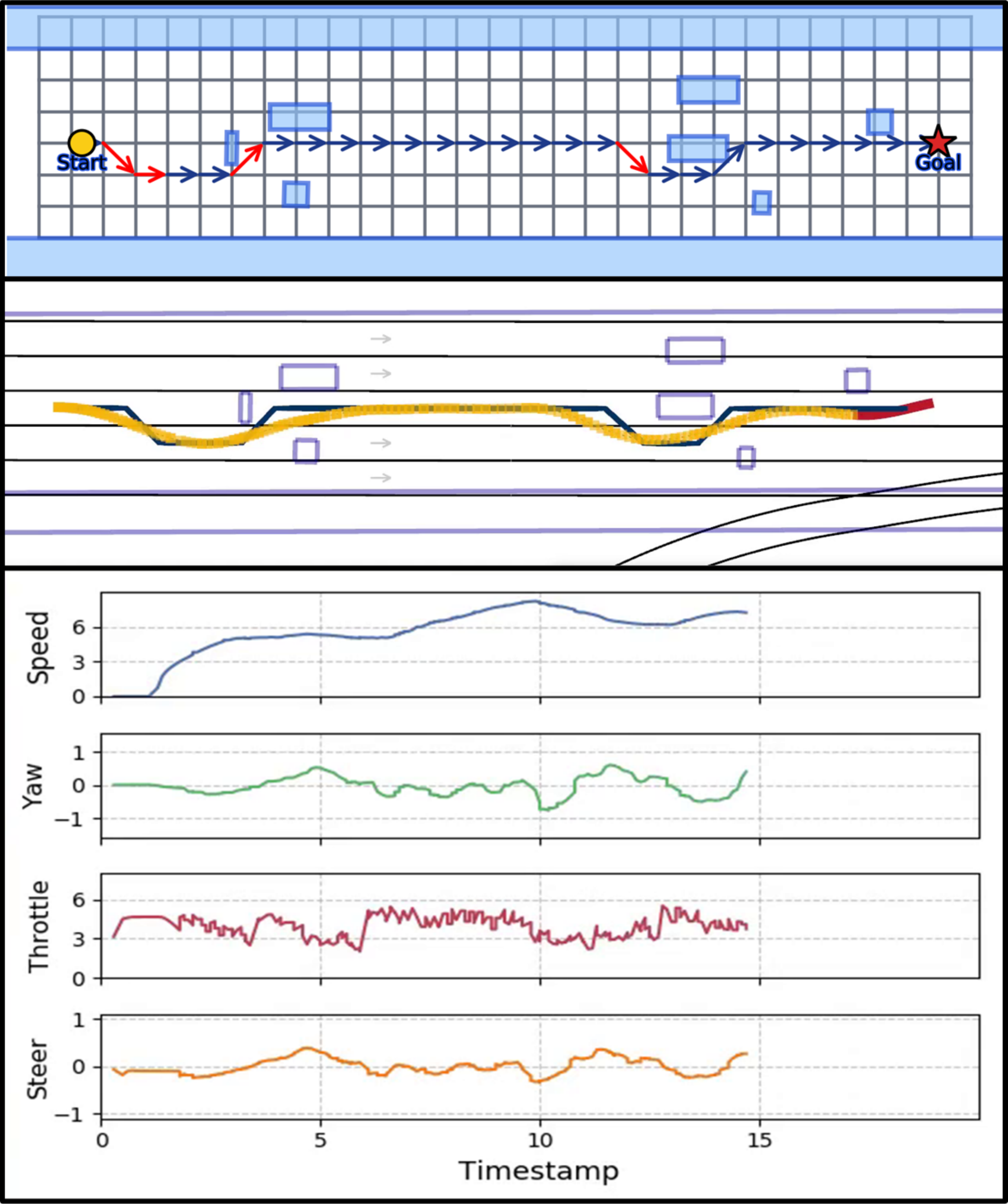}\label{fig:case1_llm}}
    \caption{Comparison of planning and control schemes in scenario~1.}
    \label{figure:scenario1}
\end{figure*}

\begin{figure}[tp]
\centering
\includegraphics[width=0.88\linewidth, height=0.4\linewidth]{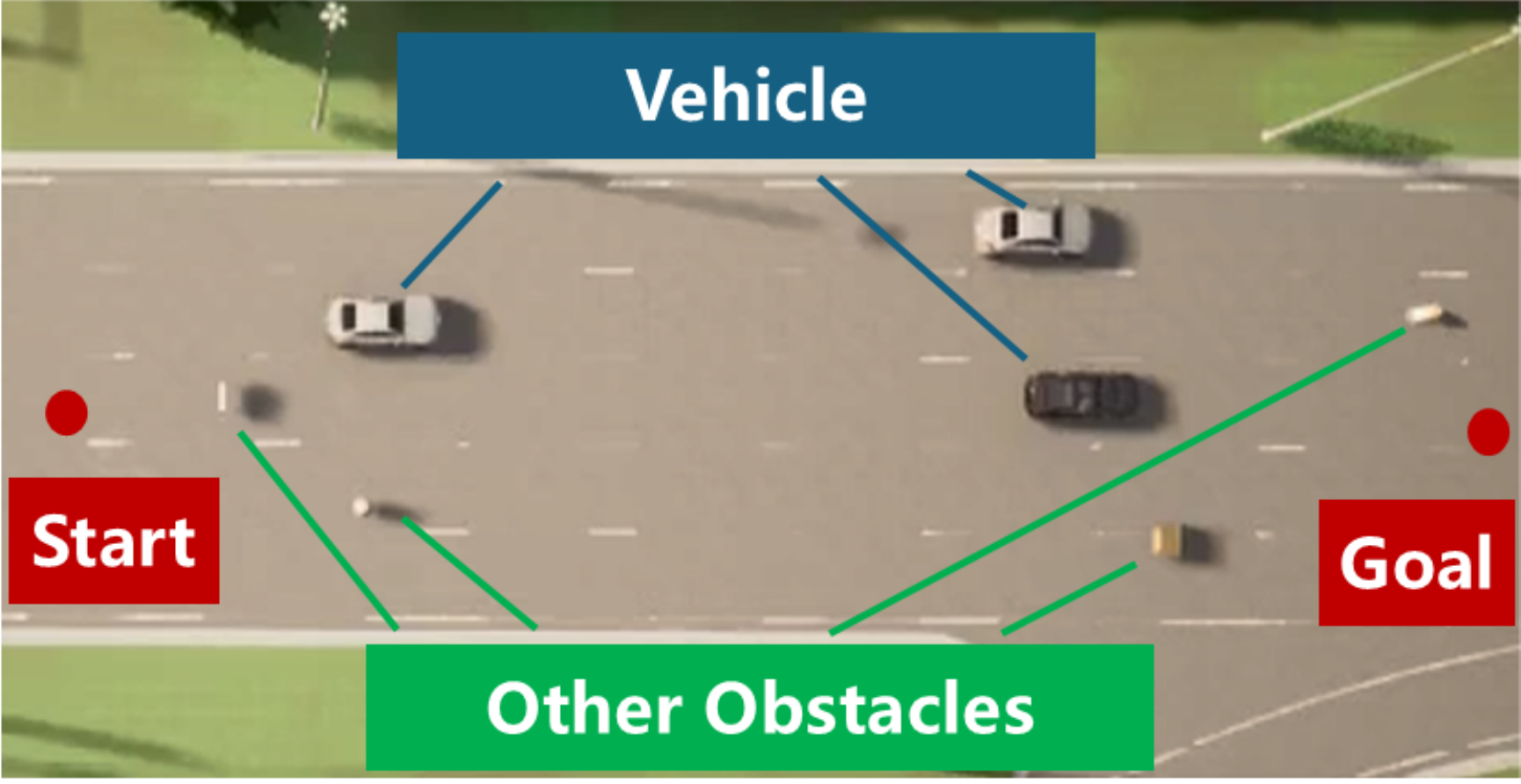}
\caption{CARLA scenario for experiment~3.}
\label{Fig:scenario}
\end{figure}

\subsection{Experiment 3: Evaluation of AFSP in CARLA scenario}
We evaluate three schemes in the CARLA simulator with a target speed of $4.2\,\mathrm{m/s}$: baseline \textbf{MPC}, an \textbf{A$^{*}$-MPC} variant combining classical A$^{*}$ planning with MPC control, and the proposed \textbf{AFSP}. Each scenario includes seven obstacles—three vehicles and four static objects—as shown in Fig.~\ref{Fig:scenario}. To account for safety margins and perception uncertainty, the vehicle inflation factor is set to $1.1$, while small static objects use enlarged radii. 

Three map types are considered: a nominal map with small inflated radii; a perturbed map with enlarged radii and a $2\,\mathrm{m}$ lateral shift in $x$; and a perturbed map with enlarged radii and shifts of $1.5\,\mathrm{m}$ in $x$ and $1\,\mathrm{m}$ in $y$. The lane width is approximately $3\,\mathrm{m}$, yielding a grid resolution of $3\,\mathrm{m}\times3\,\mathrm{m}$. Control runs at $10\,\mathrm{Hz}$ for all methods. The prediction horizon is $H=15$ with time step $\Delta t=0.2\,\text{s}$ to ensure sufficient MPC look-ahead. All schemes use the same MPC controller  \cite{han2023_rda} with fixed hyperparameters (path-following weight $w_s=0.37$, speed-following weight $w_u=0.2$); the only difference lies in the reference trajectory provided to the controller. See the attached video for details.

\subsubsection{Scenario 1: Nominal map}
As shown in Fig.~\ref{figure:scenario1}, pure MPC relies on local avoidance and often produces oscillatory control with larger lateral deviation. A$^{*}$ improves planning by providing a global route, but its grid-based design ignores kinematic feasibility; consequently, paths may deviate during execution and fail to exploit the safer lower corridor (with greater clearance). By injecting semantic guidance, \emph{our method} selects this lower corridor and produces a smoother, dynamically feasible trajectory with reduced velocity variance and maximum lateral error. Scenario~1 also includes cases where pure MPC yields the smoothest path, while A$^{*}$ may overconstrain the first decision (e.g., forcing a left turn), highlighting A$^{*}$ sensitivity and the benefit of semantic flexibility in \emph{AFSP}.

\begin{table}[h]
\caption{Quantitative comparison of experiment 3.}
\centering
\begin{tabular}{|c|c|c|c|c|c|c|}
\hline
\multirow{2}{*}{\textbf{Scen}} &
{\textbf{Method}} & {\textbf{FTime}} & {\textbf{TLenh}} & {\textbf{AvgLD}} & {\textbf{SVar}} & {\textbf{MLat}} \\
 & (Unit) & (s) & (m) & (m) & (m/s) & (m) \\
\hline
\multirow{3}{*}{1} 
 & MPC   & 17.04 & 85.96 & 2.00 & 3.01 & 6.30 \\
 & ASTAR & 17.26 & 85.91 & 1.92 & 2.13 & 6.22 \\
 & AFSP   & 14.85 & 84.21 & 1.21 & 1.56 & 3.72 \\
\hline
\multirow{3}{*}{2} 
 & MPC   & 16.80 & 85.70 & 1.89 & 2.81 & 6.26 \\
 & ASTAR & 16.63 & 83.57 & 0.98 & 1.55 & 4.17 \\
 & AFSP   & 15.02 & 83.86 & 1.01 & 1.30 & 3.23 \\
\hline
\multirow{3}{*}{3} 
 & MPC   & 16.90 & 85.54 & 1.77 & 2.70 & 6.25 \\
 & ASTAR & 16.62 & 85.49 & 1.70 & 2.46 & 5.92 \\
 & AFSP   & 14.73 & 83.72 & 1.02 & 1.48 & 3.42 \\
\hline
\end{tabular}
\label{table:quantative_result}
\end{table}

\subsubsection{Scenario 2: Shifted map with enlarged radii}
With the map shifted by $2\,\mathrm{m}$ in $x$ and with enlarged obstacle radii, A$^{*}$ performs better than pure MPC by avoiding perturbed obstacles, but parts of its route remain unsmooth and induce control fluctuations. \emph{Our method} maintains robust planning under the shift, producing smoother and kinematically consistent trajectories.

\subsubsection{Quantitative Results}
\begin{figure}[t]
    \centering
    \includegraphics[width=0.45\columnwidth]{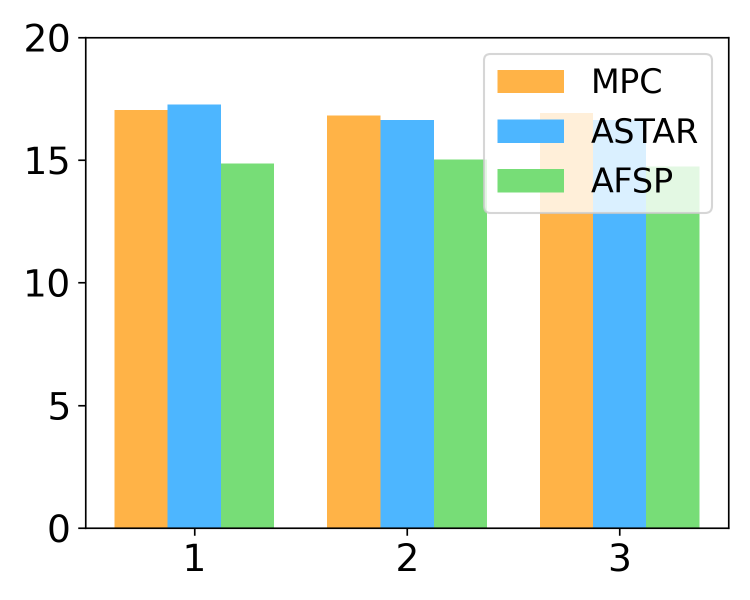}%
    \includegraphics[width=0.45\columnwidth]{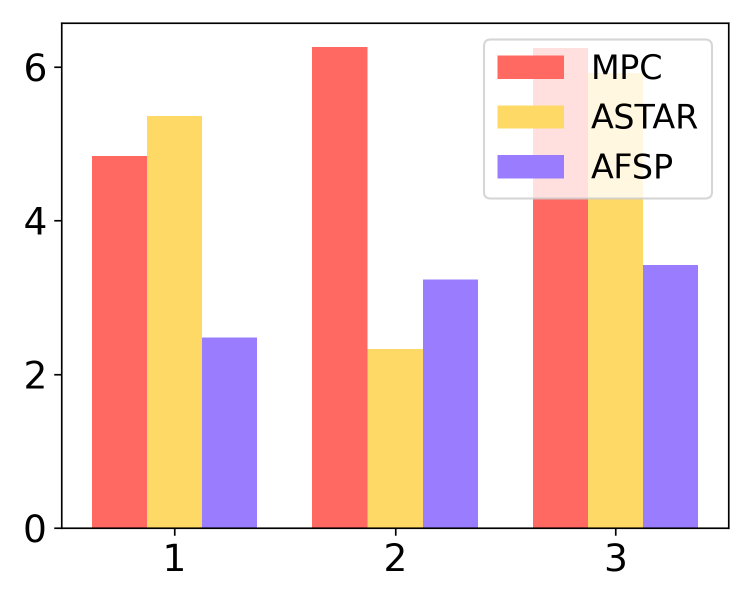}%
    \vspace{-0.07in}
    \caption{Comparison of finish time and MLat.}
    \label{fig:experiment3_timetraj}
    \vspace{-0.03in}
\end{figure}

Table~\ref{table:quantative_result} summarizes the quantitative results of Experiment~3, where each scenario is repeated 10 times to reduce randomness. Only successful runs are reported (a 100\% success rate is ensured for MPC with properly tuned parameters). We compare the proposed AFSP method with MPC and A$^{*}$ using several metrics, including Finish Time, Trajectory Length, Average Lateral Deviation, Speed Variation, and Maximum Lateral Deviation. The bar plots in Fig.~\ref{fig:experiment3_timetraj} further illustrate that AFSP achieves the shortest completion time and the smallest lateral deviation across all scenarios. On average, Finish Time is reduced by about $12\%$ compared to MPC and $11\%$ compared to A$^{*}$, while Maximum Lateral Deviation is reduced by about $45\%$ and $35\%$, respectively. For brevity, detailed visualizations of \textbf{Scenario~3} are provided in the supplementary video.

\begin{figure}[tp]
\centering
\includegraphics[width=0.9\linewidth]{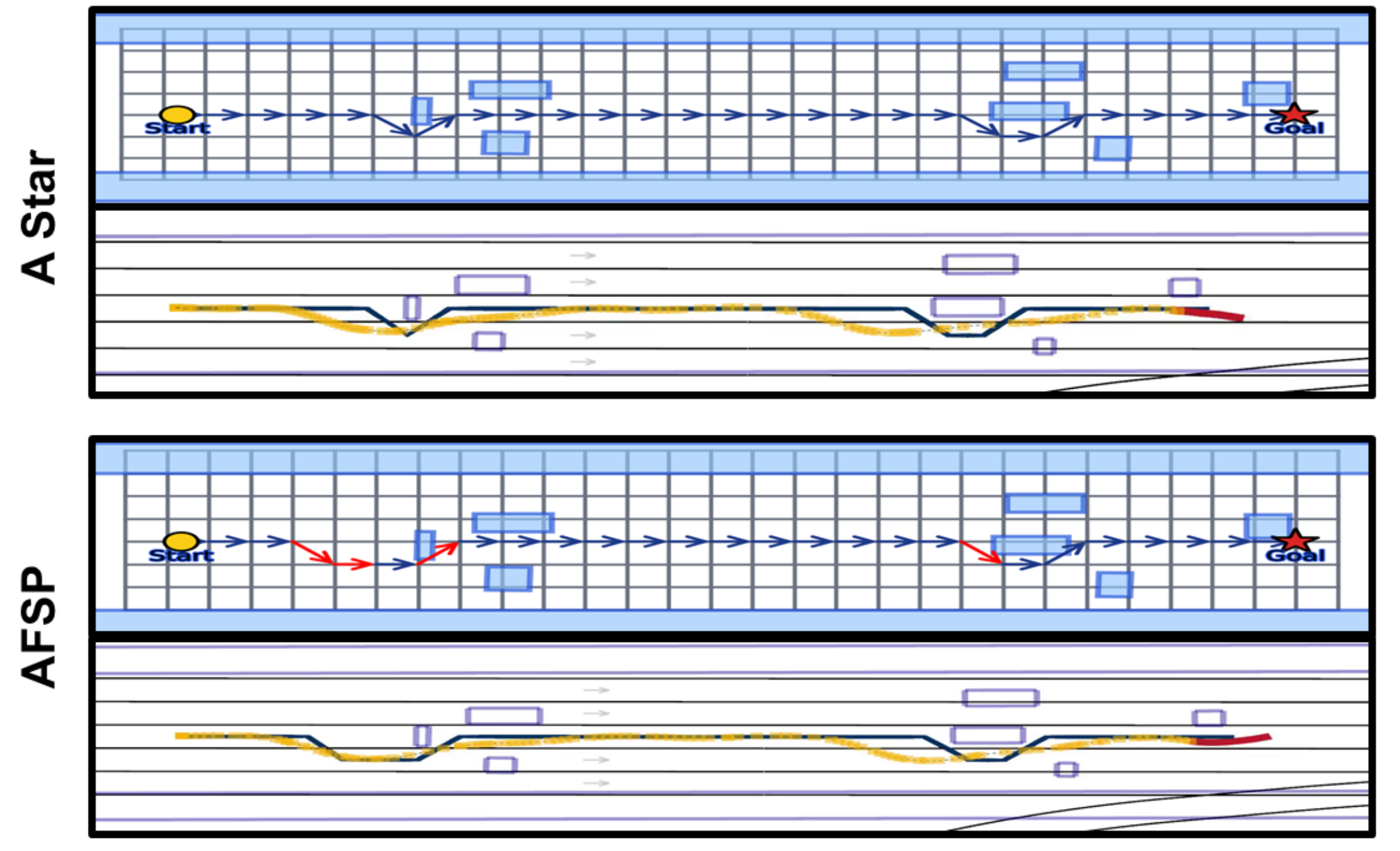}
\vspace{-0.05in}
\caption{Details in experiment3 scenario 2.}
\label{fig:exp32}
\vspace{-0.05in}
\end{figure}

\vspace{-0.05in}
\section{Conclusion}
This paper presented \textbf{Agentic Fast--Slow Planning}, a hierarchical framework that bridges perception, reasoning, planning, and control across timescales. The \textbf{Perception2Decision} module separates lightweight topology detection on the edge from semantic decision making in the cloud, reducing bandwidth while providing interpretable directives. The \textbf{Decision2Trajectory} module combines semantic-guided A$^{*}$ with agentic refinement, improving robustness and adaptability without manual tuning. Experiments in CARLA validated the effectiveness of our approach, showing safer, more efficient, and interpretable autonomous navigation compared to existing methods. Future work will focus on adaptive edge–cloud collaboration and verification of symbolic directives, moving the framework toward practical deployment.

\bibliographystyle{IEEEtran}
\bibliography{main}

\end{document}